\documentclass[wcp]{jmlr}

\usepackage{longtable}% for long tables
\usepackage{epstopdf}
\usepackage{multirow}
\usepackage{bbding}
\usepackage{array}
\usepackage{booktabs}
\jmlrvolume{80}
\jmlryear{2018}
\jmlrworkshop{ACML 2018}

\title[End-to-End Learning of Multi-scale Convolutional Neural Network for Stereo Matching]{End-to-End Learning of Multi-scale Convolutional Neural Network for Stereo Matching}
  \author{\Name{Li Zhang} \Email{sylvia\underline{ }zhang@pku.edu.cn}\\
  \Name{Quanhong Wang} \Email{ilxywang@foxmail.com}\\
  \Name{Haihua Lu} \Email{luahihua@pku.edu.cn}\\
  \Name{Yong Zhao}\thanks{} \Email{yongzhao@pkusz.edu.cn}\\
  \addr School of Electronic and Computer Engineering, Shenzhen Graduate School of Peking University, Shenzhen, China
 }

\editors{Jun Zhu and Ichiro Takeuchi}

\begin{document}

\maketitle

\begin{abstract}
Deep neural networks have shown excellent performance in stereo matching task. Recently CNN-based methods have shown that stereo matching can be formulated as a supervised learning task. However, less attention is paid on the fusion of contextual semantic information and details. To tackle this problem, we propose a network for disparity estimation based on abundant contextual details and semantic information, called Multi-scale Features Network (MSFNet). First, we design a new structure to encode rich semantic information and fine-grained details by fusing multi-scale features. And we combine the advantages of element-wise addition and concatenation, which is conducive to merge semantic information with details. Second, a guidance mechanism is introduced to guide the network to automatically focus more on the unreliable regions. Third, we formulate the consistency check as an error map, obtained by the low stage features with fine-grained details. Finally, we adopt the consistency checking between the left feature and the synthetic left feature to refine the initial disparity. Experiments on Scene Flow and KITTI 2015 benchmark demonstrated that the proposed method can achieve the state-of-the-art performance.
\end{abstract}
\begin{keywords}
stereo matching, supervised learning, multi-scale feature, guidance mechanism
\end{keywords}

\section{Introduction}
Stereo matching is an important branch of computer vision. Recent years there have seen signification process on the problem of estimating dense disparity map, which is evidenced by increasingly accuracy on the KITTI. The disparity map is obtained by finding the correspondence between a pair of stereo images. In the same way, stereo matching is also a bottleneck in the field of computer vision research, and the results are directly related to the effects of relevant technologies. Taking autonomous driving, unmanned aerial vehicle and robot as examples, stereo matching shows vital performance in non-contact measurement. Therefor the research of stereo matching is of great significance.

There are four steps in the disparity estimation pipeline. First, to measure the similarity between two images, we have to compute the matching cost of each pixel. Second, the matching cost need to be smoothed which called cost aggregation. Then the disparity prediction will be implemented to obtain an initial disparity map by searching for the minimum matching cost. Finally, there are some unstable pixels in the initial disparity map so that the initial disparity map needs to be refined.

According to the four steps, stereo matching methods can be broadly improved by matching cost computation and disparity refinement. The first convolutional neural network for stereo matching (~\cite{Z2015Stereo}) was proposed to calculate matching cost, which computed the similarity score for a pair of image patches to further determine whether they are matched. Most of patch based stereo methods focus on using CNN to extract high-level features. Following that, some methods were proposed to improve the computational efficiency (~\cite{Luo2016Efficient}) and matching accuracy (~\cite{Shaked2016Improved}). Although these methods had many significant gains compared to conventional methods in terms of both accuracy and speed, they still suffered from some limitations to find accurate corresponding points: (i) Ill-posed regions such as occlusion areas, repeated patterns, textureless regions, and reflective surfaces. (ii)Limited receptive field.

As for end-to-end networks for disparity estimation, most networks were extended a model called FlowNet (~\cite{Dosovitskiy2015FlowNet}). For instance, DispNet (~\cite{Mayer2016A}) replaced the optical estimation with disparity estimation and provided a large dataset for training. However, some problems are also remained in the process of stereo matching. Previous architectures paid less attention on extracting contextual information, they only extracted the shallow features to be the basis of the matching cost calculation. The downside of this is lacking of semantic information. Moreover, no algorithm considers the fusion of semantic information and fine-grained details while calculating the matching cost.

In this paper, we propose MSFNet, an efficient end-to-end network based on exploiting multi-scale features. There are three subnetworks: Multi-scale Features Module (MSFM), Skip Connection Hourglass Module (SCHM), and Stacked Guidance Residual Module (SGRM). Different from other methods, we focus on the combination of semantic information and details which will be applied in matching cost computation. The fusion of semantic information and details can help to obtain the similarity measure of a stereo pair. Then the features from earlier layers in the network can provide local details for computing the error map as the geometric constraints in the part of refining the initial disparity map. By fusion, we expect the subnetwork to consider meaningful sematic information and fine-gained details when performing disparity estimation over the matching cost. The fusion module is inspired by YOLOv3 (~\cite{Redmon2018YOLOv3}). The YOLOv3 model takes two feature maps from two deep layers and up-sample them by 2×, then takes two feature maps from earlier layers in the network, finally combines it with the up-sampled feature by element-wise addition. Our firstly module extends YOLOv3 by fusing multi-scale features and the abilities of element-wise addition and concatenation. The rich semantic information and fine-grained details are encoded in the subnetwork MSFM.

Then we integrate the cost aggregation and disparity estimation steps into a CNN to directly predict the disparity. This sub-network is called SCHM. Because of the outliers and depth discontinuities, we perform a disparity refinement for the initial disparity by regarding the error map between the left feature and the synthetic left feature as a guidance. The error map guides the initial disparity to improve the depth estimation performance in the mismatched regions. Our thirdly submodule is called Stacked Guidance Residual Module (SGRM).

In summary, this paper has the following six major contributions:

1)We proposed a fusion module which exploits multi-scale features to encode rich semantic information and fine-grained details.

2)A guidance mechanism is proposed to guide the network to automatically focus more on the unreliable regions learned by the consistency checking during the disparity refinement.

3)We combine the advantages of element-wise addition and concatenation, which is more conducive to the fusion of the semantic information and details.

4)Start-of-the-art accuracy is achieved on the Scene Flow dataset and the KITTI dataset.

\section{Related work}
The traditional algorithms for estimating disparity can be divided into two categories: local algorithm and global algorithm. But they rely on the hand-crafted features. These features are insufficient to provide a strong information for the disparity estimation, resulting in the low accuracy.

With the development of neural network, CNNs have been introduced to solve the problems in stereo matching. Deep learning network can extract more effective and robust features. There exists a large body of methods on stereo matching.  In contrast to hand-crafted matching cost, such as normalized cross correlation (NCC) and sum of absolute difference (SAD), CNN-based approaches measure the similarity between image patches. The first disparity estimation with convolutional networks was proposed by $\check{Z}$bontar et al.(~\cite{Z2015Stereo}), which predicted the matching degree between two patches of two images by MC-CNN and computed the stereo matching cost. In contrast to an independent computation similarity between image patches, Content-CNN (~\cite{Luo2016Efficient}) was employed to learn the probability distribution over all disparity value and capture the correlation between different disparities. Siamese network (~\cite{Bromley1993Signature}) performed feature extraction from both images and matched them using a fixed product layer. Some methods (~\cite{Shaked2016Improved}; ~\cite{Xu2017Accurate}) were employed to improve representations under the Siamese architecture. ResMatchNet (~\cite{Shaked2016Improved}) learned to measure reflective confidence for the disparity maps to improve performance in ill-posed regions. The model LW-CNN (~\cite{Park2017Look}) used a novel CNN module to learn the matching cost of a large size windows, which enabled CNN to observe large receptive field without losing details. Some methods focus on the post-processing of the disparity map. (~\cite{Guney2015Displets}) introduced a deep network to resolve matching ambiguities in reflective and textureless regions by 3D models. Moreover, DRR (~\cite{Gidaris2016Detect}) was proposed to improve the labels by detecting incorrect labels, replacing incorrect labels and refining the new labels. The SGM-Net (~\cite{Hirschmuller2005Accurate}) learned to predict SGM penalties instead of manually-tuned penalties for regularization.

\begin{figure}[tp]
\begin{center}
\includegraphics[width=6.1in,height = 2.9in]{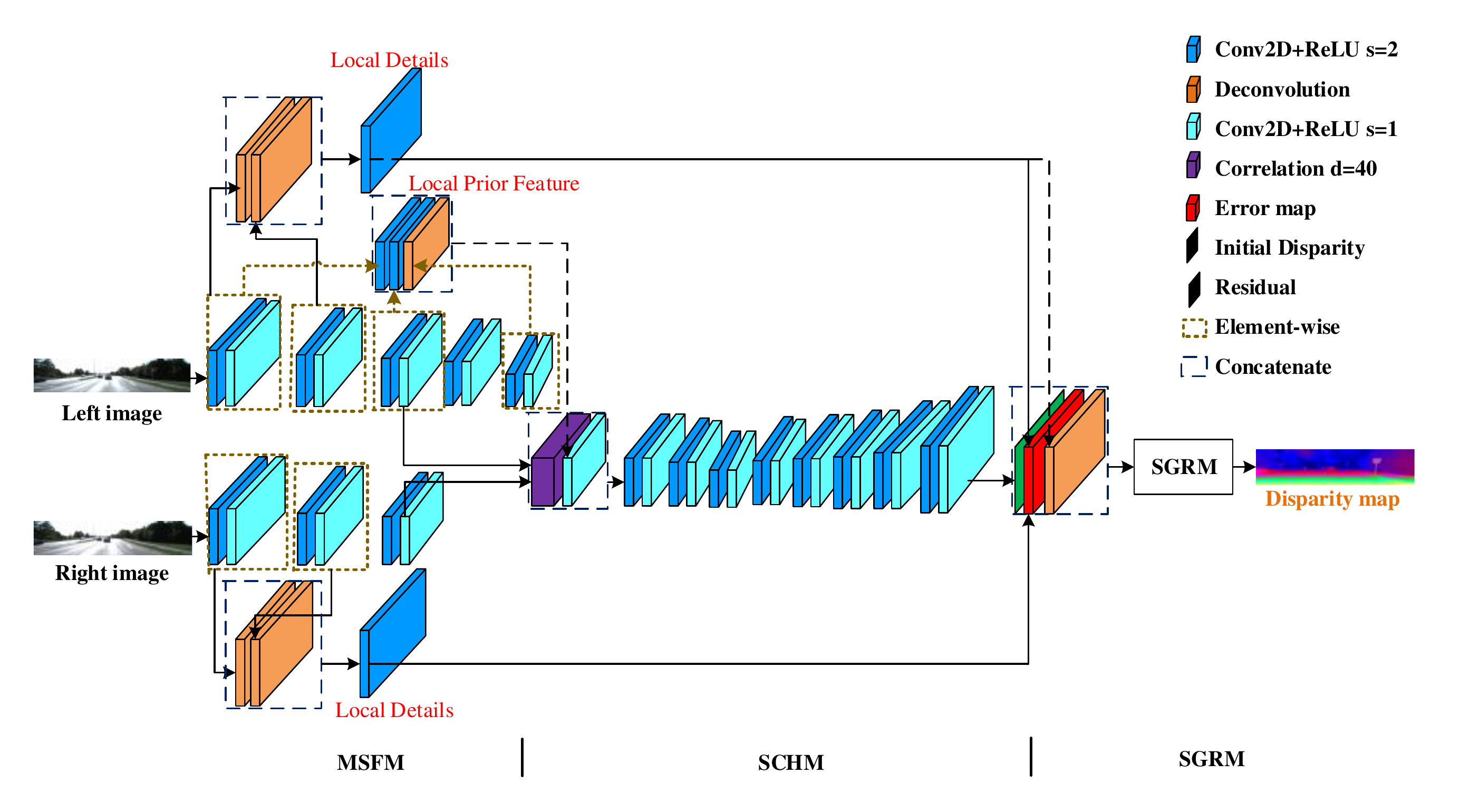}
\caption{Architecture overview of proposed MSFNet. All of the four steps for stereo matching are incorporated into a single network. Given a stereo pair, the disparity map is the output. The outputs of the first module MSFM are Local Prior Feature and Local Details. Detailed structures of SGRM is displayed in Fig.~\ref{fig:SGRM}.}
\label{fig:MSFNet}
\end{center}
\end{figure}

End-to-end deep learning methods have also been introduced to take the stereo images as input and output the final disparity map. The model FlowNet (~\cite{Dosovitskiy2015FlowNet}) was the first end-to-end CNN to estimate the optical flow. Therewith the DispNet architecture (~\cite{Mayer2016A}) was applied to the disparity estimation task and presented a synthetic stereo dataset, Scene Flow, which was a large dataset to train convolutional networks for disparity, optical flow, and scene flow estimation. A two-stage network called cascade residual learning (CRL) (~\cite{Pang2017Cascade}) extended DispNet (~\cite{Mayer2016A}) by adding a deep residual module to learn its multi-scale residuals. Then the final disparity map was formed by summing the initial disparity with the residual. In addition, GC-Net (~\cite{Kendall2017End}) learned disparity map by the geometric information and context, and introduced geometric 3D convolution and a soft-argmin layer.

Recently there has a novel network to extract contextual features. The novel model EdgeStereo (~\cite{Song2018EdgeStereo}) was the first stereo algorithm based on multi-task structure to estimate the disparity map. It was beneficial to both stereo matching task and edge detection task after the multi-task learning. Meanwhile, a pooling context pyramid was proposed to extract multi-scale contextual features in the model. But the model has some disadvantages. First, the pooling operations will lose some information although the pyramid is used to obtain multi-scale information. Second, the calculation of pyramid is large.

In other field, multi-scale fusion features have also been paid great attention. A deeply supervised fully convolutional neural network (~\cite{Xie2017Holistically}) was presented to detect edge. Multiple parallel network (~\cite{Buyssens2012Multiscale}) received some inputs at different scale. Dilation convolutions in different deep network (~\cite{Chen2014Semantic}) was introduced to aggregate multi-scale contextual information. In the paper, different from the dilation convolution (~\cite{Chen2014Semantic}), a large convolution kernel was used to obtain sufficient receptive field, and then the earlier two stacked convolution layers are fused by element-wise addition and concatenate.

To our knowledge, all of the above algorithms simply provide the features extracted from the earlier layers for the back part of the network except the model EdgeStereo(~\cite{Song2018EdgeStereo}). And there is no method to try to obtain more contextual features by merging the element-wise addition with concatenation. Out model is the first stereo method based on fusing various multi-scale features to help stereo matching.

\section{Proposed method}
We propose MSFNet, which consists of three subnetworks: Multi-scale Features Module (MSFM) for multi-scale features extraction, Skip Connection Hourglass Module (SCHM) for predicting the initial disparity map, and Stacked Guidance Residual Module (SGRM) for disparity refinement. In this section, we first introduce that how multi-layer fusion features are applied to extract dense contextual features for stereo matching. Then we discuss the modules SCHM and SGRM. The architecture is illustrated in Fig.~\ref{fig:MSFNet}.

\subsection{Multi-scale Features Module (MSFM)}
This module is designed to extract more local contextual features to compute matching cost and further produce a more accurate disparity map. Instead of using the raw images to compute matching cost, we use high-level features. In our network we learn a deep feature representation through some 2D convolution layers. The details of the subnetwork are listed in Table ~\ref{tab:ls_value1}. We form the high-level features by passing both left image and right image through these layers.

There are four methods to capture multi-scale contextual information in the semantic segmentation task. First, image pyramid (~\cite{Chen2015Attention}) extracted the features at different scales respectively. But because of the limitation of GPU memory, the model can't extend well to larger or deeper DCNNs. The second method is an encoder-decoder structure, which is also applied in our model. The structure extracts multi-scale features in the encoder and restores the resolution in the decoder part. Then the context module encodes the context of large region, while DeepLabv3 (~\cite{Chen2018Encoder}) uses dilation convolution as the context module. And spatial pyramid pooling (SPP) (~\cite{He2014Spatial}) resamples the feature of a single scale. But due to the pooling or the dilation convolution in the network, the target boundary is lost.

\begin{table}[tp]
\small
\caption{\label{tab:ls_value1} The details of parameters of MSFM. k means kernel size, s means stride, and Channels is the number of input and output channels. InpRes and OutRes are the resolution of input and output. The symbol + means element-wise addition.}
\centering  % 表居中
\begin{tabular}{|c|c|ccc|cc|c|}\hline
Type & Name & k & s & Channels & InpRes & OutRes & Input \\\hline
Conv & conv\underline{ }1a & \multirow{2}{*}{7} & \multirow{2}{*}{2} & \multirow{2}{*}{3/32} & \multirow{2}{*}{768$\times$384} & \multirow{2}{*}{384$\times$192} & image\underline{ }left \\
Conv & conv\underline{ }1b & & & & & &image\underline{ }right\\\hline
Conv & conv\underline{ }1a\underline{ }1 & \multirow{2}{*}{3} & \multirow{2}{*}{1} & \multirow{2}{*}{32/32} & \multirow{2}{*}{384$\times$192} & \multirow{2}{*}{384$\times$192} & conv\underline{ }1a \\
Conv & conv\underline{ }1b\underline{ }1 & & & & & &conv\underline{ }1b\\\hline
Conv & conv\underline{ }2a & \multirow{2}{*}{5} & \multirow{2}{*}{2} & \multirow{2}{*}{32/64} & \multirow{2}{*}{384$\times$192} & \multirow{2}{*}{192$\times$96} & conv\underline{ }1a\underline{ }1 \\
Conv & conv\underline{ }2b & & & & & &conv\underline{ }1b\underline{ }1\\\hline
Conv & conv\underline{ }2a\underline{ }1 & \multirow{2}{*}{3} & \multirow{2}{*}{1} & \multirow{2}{*}{64/64} & \multirow{2}{*}{192$\times$96} & \multirow{2}{*}{192$\times$96} & conv\underline{ }2a \\
Conv & conv\underline{ }2b\underline{ }1 & & & & & &conv\underline{ }2b\\\hline
Conv & conv\underline{ }3a & \multirow{2}{*}{5} & \multirow{2}{*}{2} & \multirow{2}{*}{64/128} & \multirow{2}{*}{192$\times$96} & \multirow{2}{*}{96$\times$48} & conv\underline{ }2a\underline{ }1 \\
Conv & conv\underline{ }3b & & & & & &conv\underline{ }2b\underline{ }1\\\hline
Conv & conv\underline{ }3a\underline{ }1 & \multirow{2}{*}{3} & \multirow{2}{*}{1} & \multirow{2}{*}{128/128} & \multirow{2}{*}{96$\times$48} & \multirow{2}{*}{96$\times$48} & conv\underline{ }3a \\
Conv & conv\underline{ }3b\underline{ }1 & & & & & &conv\underline{ }3b\\\hline
Conv & conv\underline{ }4a & \multirow{2}{*}{3} & \multirow{2}{*}{2} & \multirow{2}{*}{128/256} & \multirow{2}{*}{96$\times$48} & \multirow{2}{*}{48$\times$24} & conv\underline{ }3a\underline{ }1 \\
Conv & conv\underline{ }4b & & & & & &conv\underline{ }3b\underline{ }1\\\hline
Conv & conv\underline{ }4a\underline{ }1 & \multirow{2}{*}{3} & \multirow{2}{*}{1} & \multirow{2}{*}{256/256} & \multirow{2}{*}{48$\times$24} & \multirow{2}{*}{48$\times$24} & conv\underline{ }4a \\
Conv & conv\underline{ }4b\underline{ }1 & & & & & &conv\underline{ }4b\\\hline
Conv & conv\underline{ }5a & \multirow{2}{*}{3} & \multirow{2}{*}{2} & \multirow{2}{*}{256/512} & \multirow{2}{*}{48$\times$24} & \multirow{2}{*}{24$\times$12} & conv\underline{ }4a\underline{ }1 \\
Conv & conv\underline{ }5b & & & & & &conv\underline{ }4b\underline{ }1\\\hline
Conv & conv\underline{ }5a\underline{ }1 & \multirow{2}{*}{3} & \multirow{2}{*}{1} & \multirow{2}{*}{512/512} & \multirow{2}{*}{24$\times$12} & \multirow{2}{*}{24$\times$12} & conv\underline{ }5a \\
Conv & conv\underline{ }5b\underline{ }1 & & & & & &conv\underline{ }5b\\\hline

Add & element\underline{ }wise\underline{ }1a & - &  - & 32/32 & 384$\times$192 & 384$\times$192 & conv\underline{ }1a $+$ conv\underline{ }1a\underline{ }1 \\\hline
Add & element\underline{ }wise\underline{ }3a & - &  - & 128/128 & 96$\times$48 & 96$\times$48 & conv\underline{ }3a $+$ conv\underline{ }3a\underline{ }1 \\\hline
Add & element\underline{ }wise\underline{ }5a & - &  - & 512/512 & 24$\times$12 & 24$\times$12 & conv\underline{ }5a $+$ conv\underline{ }5a\underline{ }1 \\\hline
Conv & down\underline{ }sample\underline{ }1a & 3 &  4 & 32/32 & 384$\times$192 & 96$\times$48 & element\underline{ }wise\underline{ }1a \\\hline
Deconv & upsample\underline{ }5a & 4 &  4 & 32/32 & 24$\times$12 & 96$\times$48 & element\underline{ }wise\underline{ }5a \\\hline
\multirow{3}{*}{conv} & \multirow{3}{*}{conv\underline{ }convat1\underline{ }5\underline{ }3a} & \multirow{3}{*}{1} &  \multirow{3}{*}{1} & \multirow{3}{*}{672/64} & \multirow{3}{*}{96$\times$48} & \multirow{3}{*}{96$\times$48} & down\underline{ }sample\underline{ }1a \\
 & & & & & & & upsample\underline{ }5a  \\
 & & & & & & & element\underline{ }wise\underline{ }3a  \\\hline

Add & element\underline{ }wise\underline{ }2a & - &  - & 64/64 & 192$\times$96 & 192$\times$96 & conv\underline{ }2a $+$ conv\underline{ }2a\underline{ }1 \\\hline
Deconv & upsample\underline{ }2a & 8 &  4 & 64/32 & 192$\times$96 & 768$\times$384 & element\underline{ }wise\underline{ }2a \\\hline
Deconv & upsample\underline{ }1a & 4 &  2 & 32/32 & 384$\times$192 & 768$\times$384 & element\underline{ }wise\underline{ }1a \\\hline
\multirow{2}{*}{conv} & \multirow{2}{*}{conv\underline{ }convat\underline{ }a} & \multirow{2}{*}{1} &  \multirow{2}{*}{1} & \multirow{2}{*}{64/32} & \multirow{2}{*}{768$\times$384} & \multirow{2}{*}{768$\times$384} & upsample\underline{ }1a \\
 & & & & & & & upsample\underline{ }2a  \\\hline

Add & element\underline{ }wise\underline{ }1b & - &  - & 32/32 & 384$\times$192 & 384$\times$192 & conv\underline{ }1b $+$ conv\underline{ }1b\underline{ }1 \\\hline
Add & element\underline{ }wise\underline{ }2b & - &  - & 64/64 & 192$\times$96 & 192$\times$96 & conv\underline{ }2b $+$ conv\underline{ }2b\underline{ }1 \\\hline
Deconv & upsample\underline{ }2b & 8 &  4 & 64/32 & 192$\times$96 & 768$\times$384 & element\underline{ }wise\underline{ }2b \\\hline
Deconv & upsample\underline{ }1b & 4 &  2 & 32/32 & 384$\times$192 & 768$\times$384 & element\underline{ }wise\underline{ }1b \\\hline
\multirow{2}{*}{Conv} & \multirow{2}{*}{conv\underline{ }convat\underline{ }b} & \multirow{2}{*}{1} &  \multirow{2}{*}{1} & \multirow{2}{*}{64/32} & \multirow{2}{*}{768$\times$384} & \multirow{2}{*}{768$\times$384} &  upsample\underline{ }1b \\
 & & & & & & & upsample\underline{ }2b  \\\hline

Conv & conv\underline{ }1a\underline{ }r & \multirow{2}{*}{3} & \multirow{2}{*}{1} & \multirow{2}{*}{32/16} & \multirow{2}{*}{384$\times$192} & \multirow{2}{*}{384$\times$192} & conv\underline{ }1a \\
Conv & conv\underline{ }1b\underline{ }r & & & & & &conv\underline{ }1b\\\hline
\end{tabular}
\end{table}

Different from the above methods, some convolution layers with large convolutional kernel size are used at the beginning of the network to obtain enough receptive field. In the lower stage of a convolution neural network, the network encodes finer-grained information, but it has less semantic information because of its small receptive field. While in the high stage, the features have abundant semantic information due to large receptive field, however, the prediction is coarse. Overall, the low stage features have more finer-grained information to make more accurate spatial predictions, while the high stage features have more meaningful semantic information to give more accurate semantic predictions. Based on the advantages of the above two stages, we fuse the low stage features and the high stage features by element-wise summation and concatenation to get the contextual feature. The network can predict the ill-posed regions pixel-wise with the guidance of spatial context.

Furthermore, due to estimate disparity map corresponding the left image, we use different layers for the left image and the right image in the module because of the importance of getting rich semantic information and details of the left image. In the part of matching cost calculation, in addition to the corresponding relationship between the left and right images, what we need more is the rich contextual information of the left image. Therefore, more convolution layers are needed to extract semantic features for the left image. Five stacked convolution layers are used to encode rich semantic information and fine-grained details with multi-scale features fusion. Among them, the feature map which is extracted by the fifth stack convolution layer, will be up-sampled by 4$\times$ so that the resolution of the up-sampled feature map becomes 1/8 of the original resolution. Then first stacked convolution layer is sampled by 4$\times$ to perform concatenation with the up-sampled feature map. After the above operations, we concatenate the up-sampled feature map, the sampled feature map, and the third stacked convolution layer to obtain the final feature map, called \textbf{Local Prior Feature}, which is one of the outputs in subnetwork MSFM. For each stack of convolution layer, the two convolution layers are summed by element-wise addition. According to the Local Prior Feature, we can get more meaningful semantic information from the up-sampled feature map and fine-grained details from the earlier feature map.

Crucially, we exploit novel features through the element-wise addition and concatenation operation. The element-wise addition focuses on reusing features, and the concatenation operation benefits the discovery of new features. Unlike previous works which used concatenation directly, we first merge the first convolutional layer with the second convolutional layer using element-wise addition at each scale. It allows us to learn to incorporate the large receptive field and small receptive field. For each convolution stack, the element-wise addition is applied to get new features by reusing the feature with large convolutional kernel size and the feature with small convolutional kernel size. After three element-wise addition operations we merge these features with each other by concatenation to discovery new features with rich semantic information and fine-grained details. Then we concatenate three features with each other after they are dealt to the same scale. The efficient of the above operations in any deep learning model can be good so that it can be applied on different tasks, such as semantic segmentation, edge detection, and object detection.

In fact, different features should be used in different subtasks. The local prior feature is used to compute the matching cost because it can provide the semantic information and fine-grained details so that the matching cost can be robust. As for the disparity refinement, it needs the features with a large receptive field and enough details to obtain the more accurate error map. We perform the element-wise addition for the two features of both first and second stack convolution layers so that we can reuse the features. Then we can get two features and concatenate them to explore new features, followed by a convolution layer with the kernel of 1$\times$1. The above operations are performed simultaneously with two images. After the above operations, we can obtain \textbf{Local Details} of both the left and the right image. They will be used to compute the error map in disparity refinement step. Considering the fine-grained details, we compress the first convolution layers (conv\underline{ }1a and conv\underline{ }1b) to fewer channels to obtain conv\underline{ }1a\underline{ }r and conv\underline{ }1b\underline{ }r through a convolution layer with a kernel size of 3$\times$3. The features will be used as the inputs in the disparity refinement step. To sum up, the outputs of the first module MSFM are Local Prior Feature and Local Details. Local Prior Feature is beneficial for computing robust matching cost in second module, while Local Details are extracted to providing fine-grained details for getting the error map in third module.

\subsection{Skip Connection Hourglass Module (SCHM)}
This subnetwork is inspired by DispNetCorr1D (~\cite{Mayer2016A}) to generate a disparity map from the high-level through a typical encoder-decoder structure. In the module, a correlation layer that performs multiplicative patch comparisons between two feature maps is introduced to calculate the matching cost in feature space. There is a trade-off between accuracy and computational cost for calculating matching cost. It depends on which features we choose. More details are lost if we use high-level features to calculate matching cost, while the computational cost is high as the feature map are too large and the receptive field is too small if matching cost is calculated using low-level features. According to the middle-level features in the first module, we can compute the correlation of the two feature maps with more details and low computational cost. Calculating matching cost is an indispensable part of the stereo matching task. The maximum displacement d is 40 in the correlation1D layer.

After the calculation of the matching cost, we concatenate it with the local prior feature. We find that forming a cost volume with concatenated features improves performance. From the introduction above, it is not difficult to know that the combination of the matching cost and the local prior feature can not only provide the correspondence of the two images, but also provide the local contextual information which is rich in sematic information and the details. We can regard the concatenation as the cost aggregation, which can improve accuracy of the estimated disparity map.

The next work is to encode the above features for extracting feature at different scales. Each scale will perform the skip connection. To perform the decoder, the disparity map is estimated in the part at different scales, where skip connection is performed at each scale, the features described in encoder part are skip connected to the layers of the decoder. We apply the ‘upconvolution’ to feature maps, and concatenate it with corresponding feature maps from the encoder part of the network and an upsampled coarser disparity prediction. Then the Local Detail of left image is only skip connected to the last layer of the decoder to perform full-resolution disparity estimation. To getting a disparity map of full resolution, an extra up-convolution layer is applied to getting a feature map of full resolution. Then concatenation is performed to merge it with the multi-scale fusion feature and the predicted disparity map. It's worth noting that the final output of the decoder is an initial disparity map of full resolution, which is different from the initial disparity map in model DispNet (~\cite{Mayer2016A}), which is half of the original resolution.

\begin{figure}[tp]
\begin{center}
\includegraphics[scale=0.53]{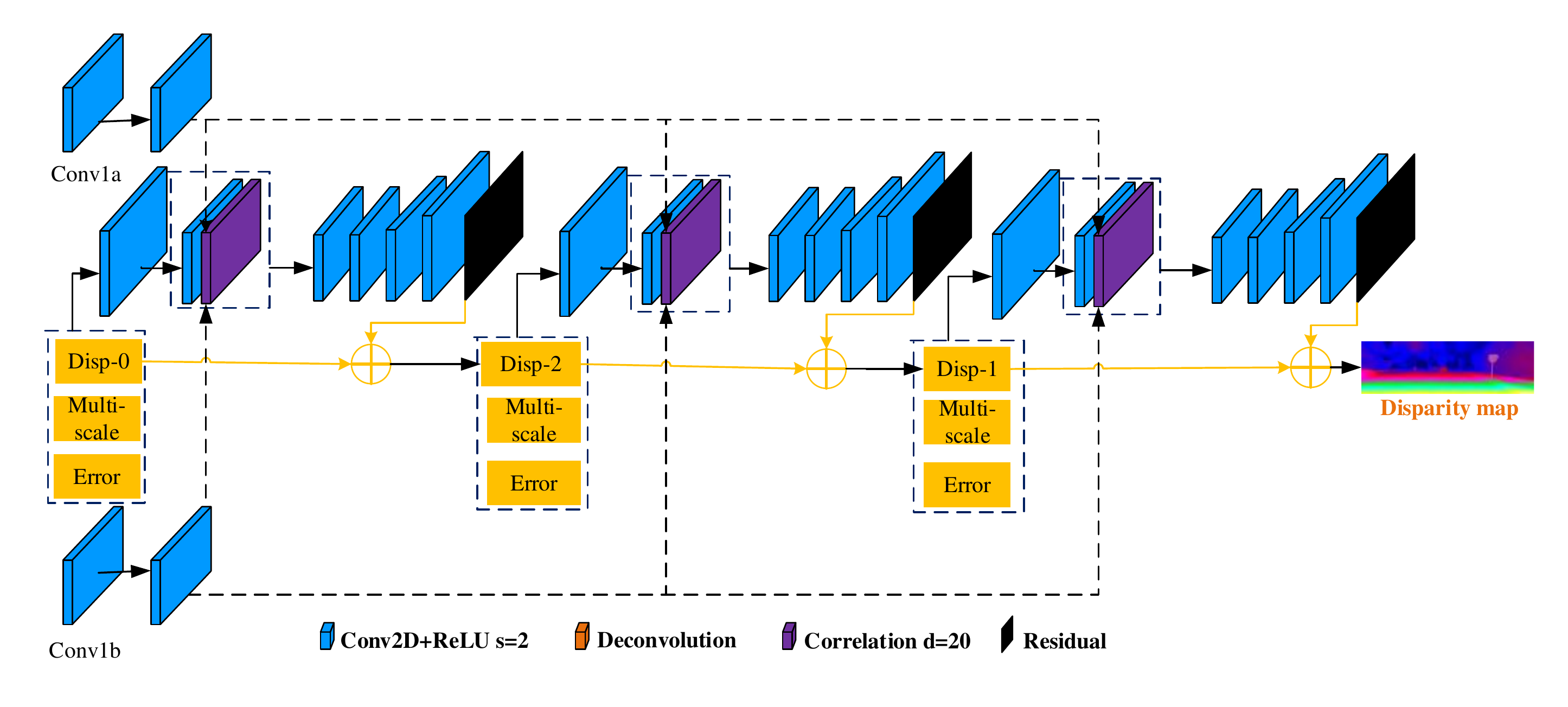}
\caption{Architecture overview of our submodule SGRM, which stacks three Guidance Residual Module. It takes the Local Details and the initial disparity as inputs to produce the residual for the initial disparity. The final disparity map is represented by the summation of the initial disparity and residual.}
\label{fig:SGRM}
\end{center}
\end{figure}
\subsection{Stacked Guidance Residual Module (SGRM)}
After the above parts, we can already get a disparity map. But there also have some problems to obtain an accurate disparity map because of the outliers and occluded regions.  The module, called SGRM, extends DispResNet (~\cite{Pang2017Cascade}) by adding a guidance mechanism. But due to the complexity, we only use three layers to encode features and three layers to decode features in the residual module. The guidance mechanism is used to guide the subnetwork to focus more on the unreliable regions learned by the consistency checking. Inspired by FlowNet (~\cite{Dosovitskiy2015FlowNet}), we use the warping operation as a transfer to obtain the synthetic feature F$_w$ (x,y) according to the high-level feature of the right image and the initial disparity, and then deal the synthetic feature F$_w$ (x,y) with the original left feature F$_L$ (x,y). In our work, we use the local detail "conv\underline{ }concat\underline{ }b" of the right image as the high-level feature. The original left feature F$_L$ (x,y) in the network is local detail "conv\underline{ }concat\underline{ }a". By the absolute difference, we expect the result to consider the error map Error$_L$ of the synthetic left feature and the raw left feature.

\begin{eqnarray}
Error_{L} = |F_{L}(x,y)-F_{w}(x,y)|  \label{eqn:speed}
\end{eqnarray}

The error map Error$_L$ will be a guidance to automatically guide the initial disparity to improve the depth estimation performance in the unreliable regions.

As of now, we can provide the initial disparity map and the reconstruction error for disparity refinement. But it is still not enough to learn the robust residuals, we need the features which can provide the details of the left image. Because of the full resolution, the best choice is the multi-scale fusion feature "conv\underline{ }concat\underline{ }a" extracted in the first module. Then we merge it with our error map and initial disparity map using concatenation. The fourth input of the subnetwork is the correlation between the feature maps of the left and right images. Given the small displacement and large receptive field, we can capture short-range but fine-grained correspondence, so we choose the features (i.e., conv\underline{ }1a\underline{ }r and conv\underline{ }1b\underline{ }r).

Finally, the initial disparity map which is the result of the second module is summated with the residual. By summation, we expect the residual to reduce the mistakes of the initial disparity and refine the occlusion regions and outliers.

A stacked refinement is presented to extract more information from the multi-scale fusion features and estimate more accurate disparity map. As the number of stack is increased, the improvement decreases. The architecture is illustrated in Fig.~\ref{fig:SGRM}.

%\subsection{Loss Function}
In model MSFNet, the whole training goal is to get an enough steady difference between the predicted disparity map and the ground-truth. We adopt the L1 loss function to measure the degree of deviation between the predicted disparity map and the ground-truth disparity map. The full-resolution disparity image, along with the other intermediate disparity images at six different scales, are supervised by the ground-truth through computing the L1 loss. h and w denote corresponding height and weight, P denote predicted disparity map, and G denote the ground-truth disparity map. Hence the L1 loss function can be denoted as:

\begin{eqnarray}
L_{1} = \frac{1}{h\times w}\sum_{(x,y)\in P}|P(x,y)-G(x,y)|  \label{eqn:speed}
\end{eqnarray}

\section{Experimental results}
This model is evaluated on two stereo datasets: Scene Flow dataset and the KITTI 2015 dataset. Experimental setting and results are presented in this section. The datasets are descripted in Section 4.1. Section 4.2 shows the experiment settings and network implementation. We show performance of the model on stereo bench-marks, then compare with other methods in Section 4.3.

\subsection{Datasets}
1.	Scene Flow: it’s a large synthetic dataset for stereo matching. There are 22390 stereo images for training and 4370 stereo images for testing in subset FlyingThings3D. Due to the exiting of large disparities in few images, we need to preprocess the dataset. By the same screening as CRL (~\cite{Pang2017Cascade}), we remove this disparity map and the corresponding stereo pair if more than 25\% of disparity values are larger than 300 in the disparity map. After the preprocessing, the dataset contains 22258 training images and 4349 testing images with height=540 and width=960. Besides, we crop the images to size of height=384 and width=768 randomly for data augmentation.

2.	KITTI 2015: it’s a real-word dataset capturing the street views from the perspective of a driving car. The dataset has 200 image pairs with ground-truth disparity for training and 200 image pairs without ground-truth disparity for testing. In order to test in training, we randomly split the 200 training images in 160 images as training dataset and 40 images as validating dataset.

\begin{table}[tp]
\small
\caption{\label{tab:ls_value2} Comparative results on the Scene Flow dataset for networks with guidance mechanism or not.}
\centering  % 表居中
\begin{tabular}{|c|c|c|}\hline
Model & $>3px(\%)$ & EPE \\\hline
Without guidance mechanism &8.93&2.15\\\hline
With guidance mechanism &5.05&1.58\\\hline
\end{tabular}
\end{table}

\begin{table}[tp]
\caption{\label{tab:ls_value3}Comparative results on the Scene Flow dataset for disparity estimation networks with different settings.}
\centering  % 表居中
\begin{tabular}{|c|c|c|c|c|}\hline   %表格6列，全部居中显示
\multicolumn{3}{|c|}{Network setting} & \multicolumn{2}{|c|}{SceneFlow}\\\hline
\multicolumn{2}{|c|}{MSFM} & \multirow{2}{*}{Local prior in SGRM} & \multirow{2}{*}{$>3px(\%)$} & \multirow{2}{*}{EPE} \\\cline{1-2}
Local Prior Feature & Local details & & & \\\hline
& \Checkmark & \Checkmark & 7.28 & 1.94\\\hline
\Checkmark & & \Checkmark & 6.51 & 2.00\\\hline
\Checkmark & \Checkmark  & \Checkmark & 5.27 & 1.61\\\hline
\Checkmark & \Checkmark  &  & 5.05 & 1.58\\\hline
\end{tabular}
\end{table}

\subsection{Detials of network and training}
The exact architectures of the network we train are shown in Fig.~\ref{fig:MSFNet}. We don't have any fully connected layers. The architecture consists of some stacked convolution layers and a ReLU nonlinearity unit after each convolution layer. As for the kernel size of the convolution layers, there are some forms: 7$\times$7, 5$\times$5, 3$\times$3. The size of convolution kernel depends on the case. Inside the stacked convolution layer, the stride of the first convolution layer is 2 and the other is 1. As for the correlation layer in MSFNet, we chose the parameters k=1, d=20, s$_1$=1, s$_2$=1 so that the maximum displacement is larger than 300 in the full resolution. As for training loss, we use the endpoint error (EPE), which is a standard error measure for stereo matching. It is the Euclidean distance between the predicted disparity and the ground truth, averaged over all pixels.

The MSFNet architecture is implemented using CAFFE. As for the optimization methods, we choose Adam (~\cite{Kingma2014Adam}) with the parameters: $\beta$$_1$ = 0.9 and $\beta$$_2$ = 0.999.  Our proposed model is end-to-end trained. Due to the limitation of GPU memory, we adopt a batch size of 2 when training the Scene Flow dataset and fine-tuning the KITTI 2015 dataset, while a batch size of 1 for testing.

For Scene Flow dataset, we start with the learning rate $\lambda$ = 0.0001 and then reduced it by a half every 100k iterations, the training was stopped at the 350k-th iteration. For KITTI dataset, we observe gradients with the learning rate $\lambda$ = 2e-5, and reduced it by a half at the 20k-th iteration and 120k-th iteration, and end at the 140k-th iteration.We use the percentage of disparities with their EPE larger than 3 pixels ($>$ 3 px), denoted as 3-pixel error.

\subsection{Results}
\textbf{Guidance mechanism.}   Guidance mechanism is used to guide the network to focus more on the unreliable regions. To demonstrate its effectiveness, the model with guidance mechanism is replaced by the model without guidance mechanism, the comparative results are shown in Table ~\ref{tab:ls_value2}. It can be observed that the model with guidance mechanism outperforms the model without guidance mechanism, with the 3-pixel-error being reduced from 8.93\% to 5.05\%, and the EPE being reduced from 2.15 to 1.58.

As listed in Table~\ref{tab:ls_value3}, we evaluate our model MSFNet with different setting, and we can verify the availability of the proposed structure according the end-point-end and the percentage of three-pixel-error on the Scene Flow test set.

\textbf{Local Prior Feature.} To demonstrate its effectiveness, we replace the local details with our local prior feature, three-pixel-error is reduced from 7.28\% to 6.51\%. This is sufficient to prove that the local prior feature is beneficial for dense disparity estimation. That is because, more semantically meaningful features and finger-gained details are obtained from the multi-scale fusion feature.

\textbf{Different features for different subtasks.} we want to demonstrate that different features should be extracted for different sub-tasks. Firstly, we use the local prior feature to compute matching cost, then use the local details to obtain the reconstruction error map. Compared with the model only with local prior feature, 3-pixel-error is reduced from 6.51\% to 5.27\%. And compared with the model only with local details, 3-pixel-error is reduced from 7.28\% to 5.27\%. As for matching cost computation, we pay more attention on the semantic information and details. But taking more concern on local details while obtaining the error map. Hence we argue that it is beneficial for the dense disparity estimation to extract the features of different concerns for different subtasks.

\textbf{Redundant information.} Given the Local Prior Feature, we try to add the feature to disparity refinement. In other words, in addition to the four inputs in the disparity refinement, we added another feature as input. We find that the 3-pixel-error of the model is 5.27\% while it’s reduced to 5.05\% for the model without the redundant information. Therefore, if there has redundant information in the network, the accuracy of disparity estimation will be reduced.

\begin{table}[tp]
\caption{\label{tab:ls_value4}Comparative results on the Scene Flow dataset for the numbers of stack.}
\centering  % 表居中
\begin{tabular}{|c|c|c|}\hline
\multirow{2}{*}{Number of stack} & \multicolumn{2}{c|}{SceneFlow}\\\cline{2-3}
 & $>3px(\%)$ &  EPE \\\hline
1&5.38&1.63\\\hline
2&5.15&1.61\\\hline
3&5.05&1.58\\\hline
\end{tabular}
\end{table}

\begin{table}[tp]
\small
\caption{\label{tab:ls_value5} Comparisons of stereo matching algorithm on Scene Flow test set.}
\centering  % 表居中
%\begin{tabular}{c|c|c|c|c|c|c|c|c|c|c}\hline
\begin{tabular}{p{1.5cm}<{\centering}|p{0.8cm}<{\centering}|p{0.7cm}<{\centering}|p{0.7cm}<{\centering}|p{0.8cm}<{\centering}|p{0.7cm}<{\centering}|p{1.2cm}<{\centering}|p{1.1cm}<{\centering}|p{0.8cm}<{\centering}|p{0.7cm}<{\centering}|p{1.5cm}<{\centering}}\hline
Metric & MC-CNN & DRR & SGM & SPS-st & CA-Net & DispNet & DispFul Net & GC-Net & CRL & \textbf{MSF-Net(ours)}\\\hline
$>3px(\%)$ &13.70&7.21&12.54&12.84&5.62&9.67&8.61&7.20&6.20&5.05\\\hline
EPE &3.79& - &4.50&3.98& - &1.84&1.75& - &1.32&1.58\\\hline
\end{tabular}
\end{table}

\begin{figure}[t]
\begin{center}
\includegraphics[scale=0.5]{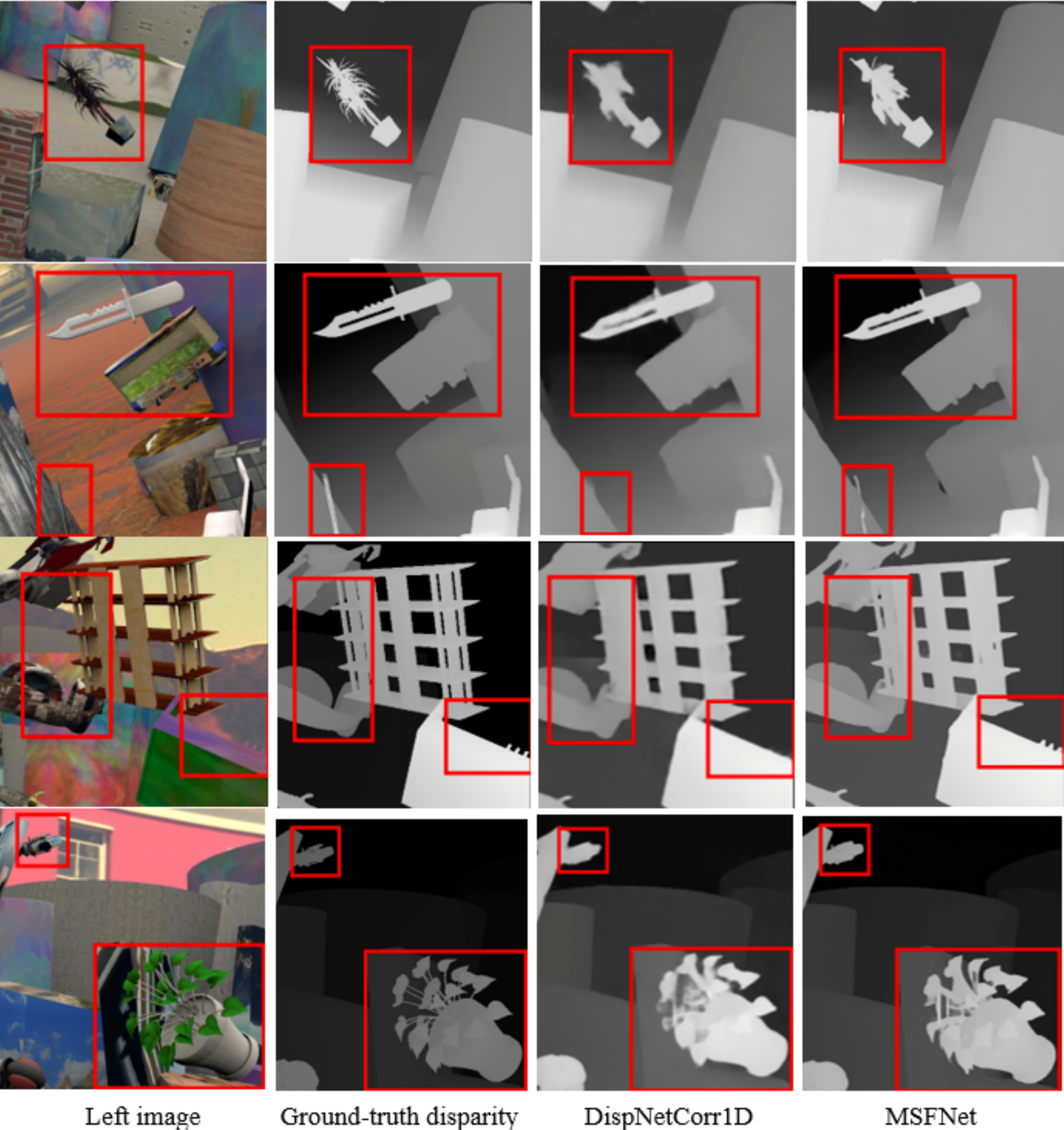}
\caption{Comparisons of different structures for stereo matching on Scene Flow dataset.}
\label{fig:sceneflow}
\end{center}
\end{figure}

To further demonstrate the effectiveness of stacked refinement, the disparity estimation results are shown in Table~\ref{tab:ls_value4}. Then we chose some non-end-to-end methods with public codes, including SGM (~\cite{Hirschmuller2005Accurate}), SPS-St (~\cite{Yamaguchi2014Efficient}), MC-CNN-fst (~\cite{Z2015Stereo}) and DRR (~\cite{Gidaris2016Detect}). And we chose the most advanced end-to-end models with public codes or results in the paper, including DispNetC (~\cite{Mayer2016A}), DispFulNet (~\cite{Pang2017Cascade}), CRL (~\cite{Pang2017Cascade}), GC-Net (~\cite{Kendall2017End}) and CA-Net (~\cite{Yu2018Deep}). The comparisons are presented in Table~\ref{tab:ls_value5}. We can see that our end-to-end model achieves the best performance in terms of two evaluation metrics. As can be seen, the performance of model is improved and we also give visual demonstrations shown in Fig.~\ref{fig:sceneflow}.

\begin{figure}[t]
\begin{center}
\includegraphics[scale=0.3]{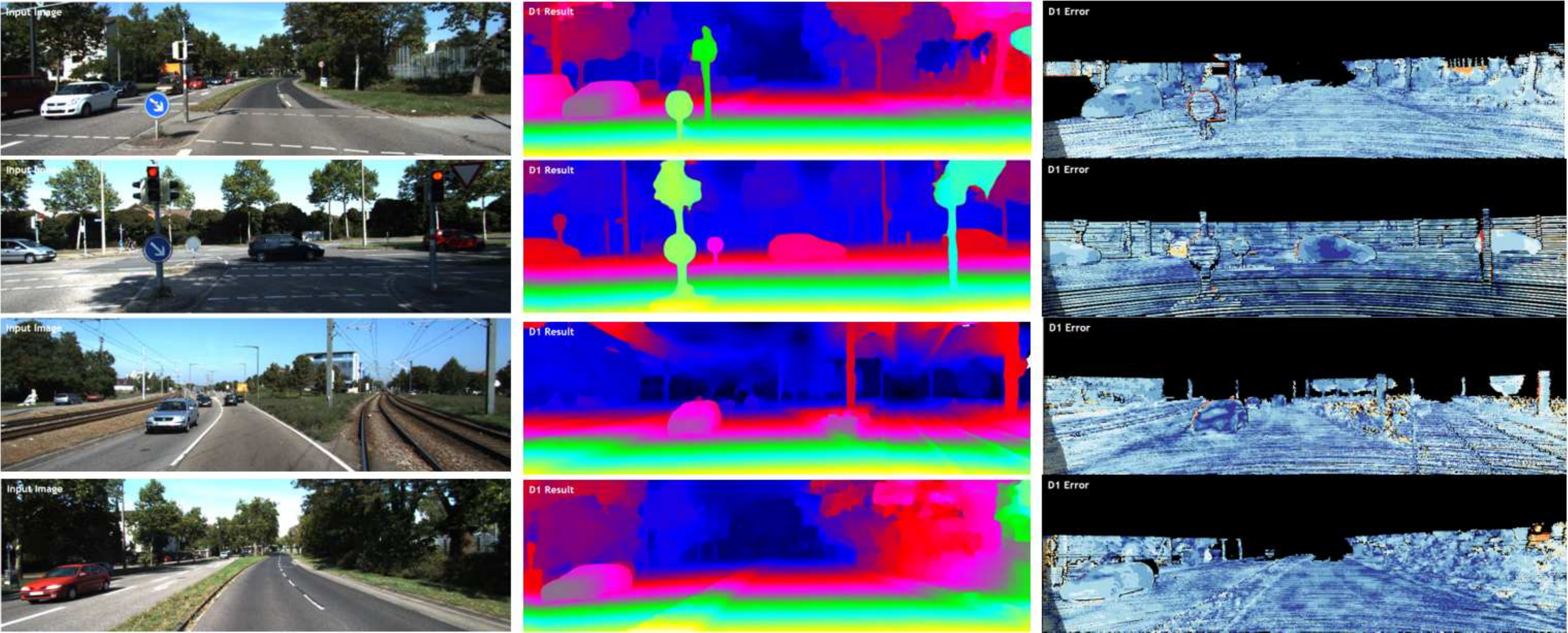}
\caption{Qualitative results on KITTI 2015 test set. From left: left stereo input image, disparity prediction, error map.}
\label{fig:kitti}
\end{center}
\end{figure}

According to the online leaderboard, as shown in Table~\ref{tab:ls_value6}. In the table, “bg” means the percentage of outliers averaged only over background regions, and “fg” means the percentage of outliers averaged only over foreground regions. The overall three-pixel-error for the proposed MSFNet is 3.15\%. It’s slightly inferior than GC-Net (2.87\%) and EdgeStereo (2.99\%). However MSFNet outperforms GC-Net obviously on Scene Flow test set. GC-Net used 3D convolutions upon the matching costs to incorporate contextual information and introduced a differentiable “softargmin” operation to regress the disparity. But high-dimensional feature volume based 3D convolution is computationally expensive. EdgeStereo is a CNN based multi-task learning network by adding an edge detection network $HED$$_\beta$. The reason why EdgeStereo method is accurate than our model is that the edge information can provide more details of object. In other words, detection task is as an auxiliary task in EdgeStereo method. At the same time, however, with the addition of edge detection network, the complexity and computational cost of the model are also greatly increased. In Fig.~\ref{fig:kitti} we show qualitative results of our method on KITTI 2015.

\begin{table}[tp]
\caption{\label{tab:ls_value6} Results on the KITTI 2015 stereo online leaderboard.}
\centering  % 表居中
\begin{tabular}{c|ccc|ccc}\hline
\multirow{2}{*}{Model} & \multicolumn{3}{c|}{All pixels(\%)} & \multicolumn{3}{c}{Non-Occluded pixels(\%)} \\
& D1-bg & D1-fg & D1-all & D1-bg & D1-fg & D1-all\\\hline
GC-Net &2.21&6.16&2.87&2.02&3.12&2.45\\
EdgeStereo &2.70&4.48&2.99&2.50&3.87&2.72\\
DRR &2.58&6.04&3.16&2.34&4.87&2.76\\
L-ResMatch &2.72&6.95&3.42&2.35&5.74&2.91\\
Displet v2 &3.00&5.56&3.43&2.73&4.95&3.09\\
PBCP &2.58&8.74&3.61&2.27&7.71&3.17\\
SGM-Net &2.66&8.64&3.66&2.23&7.44&3.09\\
MC-CNN-arct &2.89&8.88&3.88&2.48&7.64&4.05\\
DispNetC &4.32&4.41&4.34&4.11&3.72&4.05\\\hline
\textbf{MSF-Net(ours)} &2.70&5.36&3.15&2.52&4.94&2.92\\\hline
\end{tabular}
\end{table}

\section{Conclusions}
In our work, we propose a new end-to-end network architecture for stereo matching based on fusion multi-scale features. Moreover, we introduce a guidance mechanism and multi-scale fusion feature and prove their validity. Multi-scale fusion features and the guidance mechanism can be widely used in various tasks, such as semantic segmentation, object detection and edge detection. And the guidance mechanism can regard as a geometric constraint to guide network to automatically focus more on the unreliable regions. Our proposed model “MSFNet” consists of three parts: Multi-scale Features Module (MSFM) employs the multi-scale features to encode rich contextual information and local prior feature which is rich in more meaningful semantic information and fine-grained details. Then Skip Connection Hourglass Module (SCHM) regards the multi-scale fusion features as a prior information to estimate disparity map. Finally, Stacked Guidance Residual Module (SGRM) considers the above intermediate results of the former parts as the basis of estimating residual, and improves the accuracy according to the guidance mechanism. Our experimental results show that the model sets a new state-of-the-art performance on the Scene Flow dataset and the KITTI 2015 dataset.

%\acks{Acknowledgements should go at the end, before appendices and references.}

%\bibliographystyle{plain}
\bibliography{acml18}

\end{document}